%% file: LearningDynamics_WACV2020.tex
\documentclass[10pt,twocolumn,letterpaper]{article}

\usepackage{wacv}
\usepackage{times}
\usepackage{epsfig}
\usepackage{graphicx}
\usepackage{amsmath}
\usepackage{amssymb}

\input{macros.tex}

\usepackage[pagebackref=true,breaklinks=true,letterpaper=true,colorlinks,bookmarks=false]{hyperref}

\wacvfinalcopy 

\def\wacvPaperID{210} 

\newcommand\blfootnote[1]{%
  \begingroup
  \renewcommand\thefootnote{}\footnote{#1}%
  \addtocounter{footnote}{-1}%
  \endgroup
}

\ifwacvfinal\fi
\begin{document}

\title{Predicting the Physical Dynamics of Unseen 3D Objects}

\author{Davis Rempe \qquad Srinath Sridhar \qquad He Wang \qquad Leonidas J. Guibas \\ 
Stanford University }

\maketitle
\thispagestyle{empty}
\input{content/00_abstract.tex}
\input{content/01_introduction.tex}
\input{content/02_related_work.tex}
\input{content/03_problem.tex}
\input{content/04_data.tex}
\input{content/05_method.tex}
\input{content/06_results.tex}
\input{content/07_limitations.tex}
\input{content/08_conclusion.tex}

\parahead{Acknowledgments} 
This work was supported by a Vannevar Bush Faculty Fellowship, the AWS Machine Learning Awards Program, the Samsung GRO program, and the Toyota-Stanford Center for AI Research. Toyota Research Institute (``TRI'') provided funds to assist the authors with their research but this article solely reflects the opinions and conclusions of its authors and not TRI or any other Toyota entity.

{\small
\bibliographystyle{ieee}
\bibliography{references}
}

\end{document}

%% file: macros.tex
\usepackage[dvipsnames]{xcolor}
\usepackage{float}
\usepackage{multirow}
\usepackage{booktabs}

\newcommand{\parahead}[1]{\par\textbf{#1}:\ }

\usepackage{bm}
\newcommand{\vect}[1]{\boldsymbol{\mathbf{#1}}}

\newcommand{\filluptopage}[1]{%
  \clearpage
  \loop\ifnum\value{page}<#1\relax
    \null\clearpage
  \repeat
  \loop\ifnum\value{page}=#1\relax
    \null\clearpage
  \repeat
}


%% file: content/00_abstract.tex
\begin{abstract}
  Machines that can predict the effect of physical interactions on the dynamics of previously unseen object instances are important for creating better robots and interactive virtual worlds. In this work, we focus on predicting the dynamics of 3D objects on a plane that have just been subjected to an impulsive force. In particular, we predict the changes in state---3D position, rotation, velocities, and stability. Different from previous work, our approach can generalize dynamics predictions to object shapes and initial conditions that were unseen during training. Our method takes the 3D object's shape as a point cloud and its initial linear and angular velocities as input. We extract shape features and use a recurrent neural network to predict the full change in state at each time step. Our model can support training with data from both a physics engine or the real world. Experiments show that we can accurately predict the changes in state for unseen object geometries and initial conditions.
\end{abstract}

\blfootnote{Contact: \texttt{drempe@stanford.edu}}
\blfootnote{Project Webpage: \url{geometry.stanford.edu/projects/learningdynamicsWACV2020}{}}

%% file: content/01_introduction.tex
\section{Introduction}
\label{sec:intro}
We study the problem of learning to predict the physical dynamics of 3D rigid bodies with previously unseen shapes. The ability to interact with, manipulate, and predict the dynamics of objects encountered for the first time would allow for better home robots and virtual or augmented worlds. Humans can intuitively understand and predict the effect of physical interactions on novel object instances (\eg, putting a peg into a hole, catching a ball) even from a young age~\cite{baillargeon1990top,leslie1982perception}. Endowing machines with the same capability is a challenging and unsolved problem.

Learned dynamics has numerous advantages over traditional simulation. Although the 3D dynamics of objects can be approximated by simulating well-studied physical laws, this requires exact specification of properties and system parameters (\eg, mass, moment of inertia, friction) which may be challenging to estimate, especially from visual data. Additionally, many physical phenomena such as planar pushing~\cite{yu2016push} do not have accurate analytical models. Learning dynamics directly from data, however, can implicitly model system properties and capture subtleties in real-world physics. This allows for improved \emph{accuracy} in future predictions. 
Using neural networks for learning additionally offers \emph{differentiability} which is useful for gradient-based optimization and creates flexible models that can trade off speed and accuracy. There has been increased recent interest in predicting object dynamics, but a number of limitations remain. First, most prior work lacks the ability to generalize to shapes unseen during training time~\cite{byravan2017se3}, or lacks scalability~\cite{li2019particledynamics,mrowca2018flexible}.
Second, many methods are limited to 2D objects and environments ~\cite{battaglia2016interactionnets,chang2017compositional,fragkiadaki2016visualbilliards,watters2017vin} and cannot generalize well to 3D objects. 
Lastly, many methods use images as input~\cite{mottaghi2016if,mottaghi2016newton,finn2016videoprediction,agrawal2016poke} which 
provide only partial shape information possibly limiting the accuracy of forward prediction compared to full 3D input~\cite{rempe2019finalstate}, and may entangle variations in object appearance with physical motion.

Our goal is to learn to predict the dynamics of objects from their 3D shape, and generalize these predictions to previously unseen object geometries. To this end, we focus on the problem of accurately predicting, at each fixed time step, the change in \emph{object state}, \ie, its 3D position,  rotation, linear and angular velocities, and stability.
We assume that the object initially rests on a plane and has just been subjected to an impulsive force resulting in an initial velocity.
Consequently, the object continues to move along the plane resulting in one of two possible outcomes: (1)~friction eventually brings it to a rest, or (2)~the object topples onto the plane (see Figure~\ref{fig:problem}).
This problem formulation is surprisingly challenging since object motion depends non-linearly on factors such as its moment of inertia, contact surface shape, the initial velocity, coefficient of restitution, and surface friction. Objects sliding on the plane could move in 3D resulting in wobbling motion. 
Excessive initial velocities could destablize objects leading to toppling. Learning these subtleties in a generalizable manner requires a deep understanding of the connection between object shape, mass, and dynamics.
At the same time, this problem formulation has many practical applications, for instance, in robotic pushing of objects, and is a strong foundation for developing methods to predict more complex physical dynamics.

To solve this problem, we present a neural network model that takes the object shape and its initial linear and angular velocities as input, and predicts the change in object state---3D position, rotation, velocities, and stability (13 parameters)---at each time step. We use a 3D point cloud to represent the shape of the object since it is compact and decouples object motion from appearance variation, and unlike other 3D representations can be easily captured in the real world with commodity depth sensors.
To train this network, we simulate the physics of a large number of household object shapes from the ShapeNet repository~\cite{chang2015shapenet}. Our network learns to extract salient shape features from these examples. This allows it to learn to make accurate predictions not just for initial velocities and object shapes seen during training, but also for unseen objects in novel shape categories with new initial velocities.

We present extensive experiments that demonstrate our method's ability to learn physical dynamics that generalize to unseen 3D object shapes and initial velocities, and adapt to unknown frictions at test time. Experiments show the advantage of our object-centric formulation compared to a recent approach~\cite{mrowca2018flexible}. Finally, we show the ability to learn dynamics directly from real-world motion capture observations, demonstrating the flexibility of our method.

%% file: content/02_related_work.tex
\section{Related Work}
\label{sec:relwork}
The problem of \emph{learned physical understanding} has been approached in many ways, resulting in multiple formulations and ideas of what it means to \emph{understand} physics. Some work answers questions related to physical aspects of a scene~\cite{battaglia2013simunderstanding,zhang2016blocks,li2016fall,li2017stability,lerer2016fbtowers,mirza2016unsupervised}, while others learn to infer physical properties of objects from video frames~\cite{wu2015galileo,wu2016phys101,wu2017deanimation,monszpart2016SMASH}, image and 3D object information~\cite{liu2018ppd}, or intuitive physics~\cite{kubricht2017intuitive,smith2019intphys}. We limit our discussion to work most closely related to ours, \ie, learning to predict dynamics.

\parahead{Forward Dynamics Prediction}
Many methods that attempt direct forward prediction of object dynamics take the current state of objects in a scene, the state of the environment, and any external forces as input and predict the state of objects at future times. Forward prediction is a desirable approach as it can be used for action planning~\cite{hamrick2016decision} and animation~\cite{grzeszczuk2000neuroanimator}. Multiple methods have shown success in 2D settings~\cite{fraccaro2017disentangled}. \cite{fragkiadaki2016visualbilliards} uses raw visual input centered around a ball on a table to predict multiple future positions. The \emph{neural physics engine}~\cite{chang2017compositional} and \emph{interaction network}~\cite{battaglia2016interactionnets} explicitly model relationships in a scene to accurately predict the outcome of complex interactions like collisions between balls. \cite{watters2017vin} builds on~\cite{battaglia2016interactionnets} by adding a front-end perception module to learn a state representation for objects. These 2D methods exhibit believable results, but are limited to simple primitive objects.
Learned forward dynamics prediction can be useful for physical inference and system property prediction~\cite{byravan2018se3pose,zheng2018percpred,steenkiste2018relationalem,kipf2018neural}. A differentiable physics engine would facilitate this and has been demonstrated previously~\cite{peres2018diffengine,schenck2018spnets,hu2019chainqueen}.
However, it is unclear if the accuracy of these methods is sufficient for real-world applications.

\parahead{Dynamics in Images \& Videos}
Many methods for 3D dynamics prediction operate on RGB images or video frames~\cite{ye2018interpretable,riochet2018intphys,ehrhardt2017mechanics,ehrhardt2017longterm,stewart2017label,ehrhardt2018visualobs,janner2019ooprediction}. \cite{mottaghi2016newton} and \cite{mottaghi2016if} introduce multiple algorithms to infer future 3D translations and velocities of objects given a single static RGB image. Some methods directly predict pixels of future frames conditioned on actions~\cite{oh2015atari}. \cite{finn2016videoprediction} infers future video frames involving robotic pushing conditioned on the parameters of the push and uses this prediction to plan actions~\cite{finn2017planning}. \cite{ajay2018pushing} study the case of planar pushing and bouncing. In a similar vein, \cite{agrawal2016poke} uses video of a robot poking objects to implicitly predict object motion and perform action planning with the same robotic arm. Many of these methods focus on real-world settings, but do not use 3D information and possibly entangle object appearance with physical properties.
\begin{figure*}[!ht]
\begin{center}
   \includegraphics[width=\linewidth]{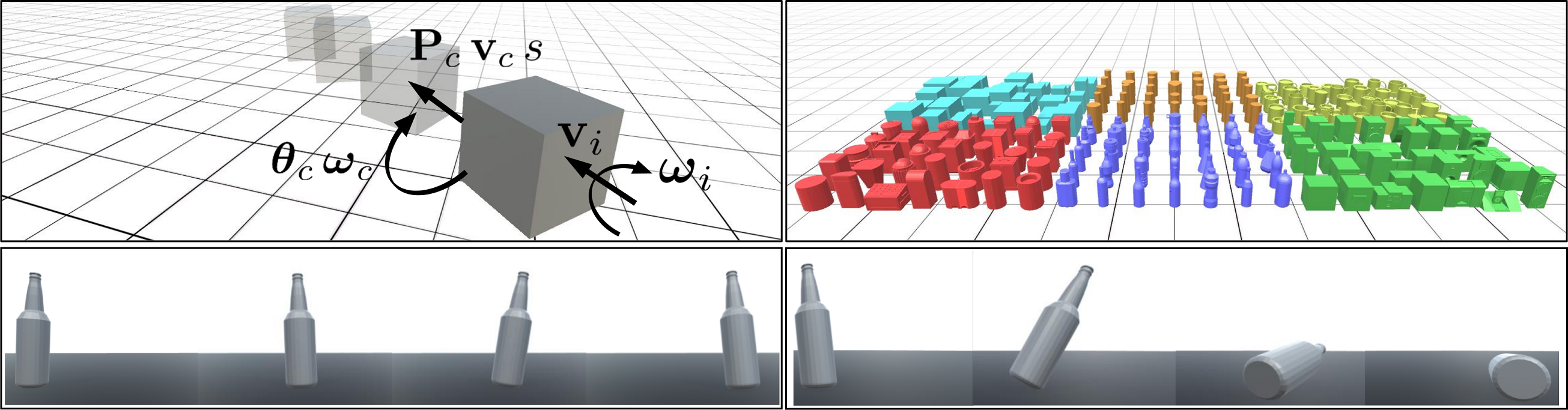}
\end{center}
   \caption{
   We study the problem of predicting the 3D dynamics of an object with linear and angular velocities, $\vect{v}_i$ and $\vect{\omega}_i$ (top left).
   Our goal is to predict, at each fixed time step, the change in object state $\vect{T}_c$, \ie, change in 3D position ($\vect{P}_c$), rotation ($\vect{\theta}_c$), linear and angular velocities ($\vect{v}_c, \vect{\omega}_c$), and stability ($s$).
   Our method can predict the dynamics of a variety of different shapes (top right) and generalizes to previously unseen object shapes and initial velocities.
   Our problem formulation presents many challenges including the unpredictable 3D motion caused due to \emph{wobbling} of objects under motion (bottom left), and object \emph{toppling} due to destabilization (bottom right).
   }
\label{fig:problem}
\end{figure*}

\parahead{3D Physical Dynamics}
Recent work has taken initial steps towards more general 3D settings~\cite{wang2018physnet,li2019particledynamics,li2019propnet,xu2019densephysnet}. Our method is most similar to \cite{byravan2017se3} who use a series of depth images to identify rigid objects and predict point-wise transformations one step into the future, conditioned on an action. However, they do not show generalization to unseen objects. Other work extends ideas introduced in 2D by using variations of graph networks. \cite{sanchez2018graphnet} decomposes systems containing connected rigid parts into a graph network of bodies and joints to make single-timestep forward predictions. The hierarchical relation network (HRN)~\cite{mrowca2018flexible} breaks rigid and deformable objects into a hierarchical graph of particles to learn particle relationships and dynamics from example simulations. Though HRN is robust to novel objects, it requires detailed per-particle supervision and results are shown only on simulated data.

%% file: content/03_problem.tex
\section{Problem Formulation}
\label{sec:problem}
We investigate the problem of predicting the 3D dynamics of a rigid object moving along a plane with an initial velocity resulting from an impulsive force. We assume the following \textbf{inputs}: (1)~the shape of the object in the form of a point cloud ($\mathbf{O} \in \mathbb{R}^{N\times 3}$), and (2)~the initial linear and angular velocities.
We further assume that the object moves on a plane under standard gravity (see Figure~\ref{fig:formulation_detail}), the friction coefficient and the restitution are constant, the object has a uniform density, and that the object eventually comes to rest due to friction and the absence of external forces.

Our goal is to accurately predict the change in state $\mathbf{T}_c^t$ (we omit the superscript for brevity) of the object at each fixed time step $t$ until it comes to rest or topples over. Specifically, we predict the change in 3D position ($\vect{P}_c \in \mathbb{R}^3$), rotation ($\vect{\theta}_c \in \mathbb{R}^3$ where $|\vect{\theta}_c|$ denotes the angle, and $\bar{\vect{\theta}}_c$ the axis), linear velocity ($\vect{v}_c \in \mathbb{R}^3$), angular velocity ($\vect{\omega}_c \in \mathbb{R}^3$), and binary stability state ($s \in \{0, 1\}$) for a total of 13 parameters. The stability state indicates whether the object has toppled over.
We continue to predict object state even after toppling, but the motion of the object after toppling is stochastic in the real world making it hard to predict accurately. For this reason, we focus evaluation of model-predicted trajectories (see Section~\ref{sec:results}) on shape generalization for sliding examples without toppling. As shown in Figure~\ref{fig:problem}, we model the 3D motion along a plane but 3D object motion is unrestricted otherwise.
The object can and does exhibit complex wobbling motion or topples over when destabilized. Unobserved quantities (\eg, mass, volume, moment of inertia) additionally contribute to the difficulty of this problem.
Such a formulation has numerous practical applications, for instance a robotic arm pushing objects on a desk to reach a goal state, and uses data which lends itself to real-world use. We use a point cloud to encode object geometry since it only depends on the surface geometry, making it agnostic to appearance, and can be readily captured in a real-world setting through commodity depth sensors. Additionally, the initial velocities of the object can be estimated from the point clouds and video.

%% file: content/04_data.tex
\section{Data Simulation}
\label{sec:data}
We use 3D simulation data from the Bullet physics engine~\cite{bullet} in Unity~\cite{unity} for our task.
However, our method can also be trained on real-world data provided ground truth shape and initial velocities are available. In fact, we show results on real motion capture data in Section~\ref{sec:real_results}.

\begin{figure}[!ht]
\begin{center}
   \includegraphics[width=\linewidth]{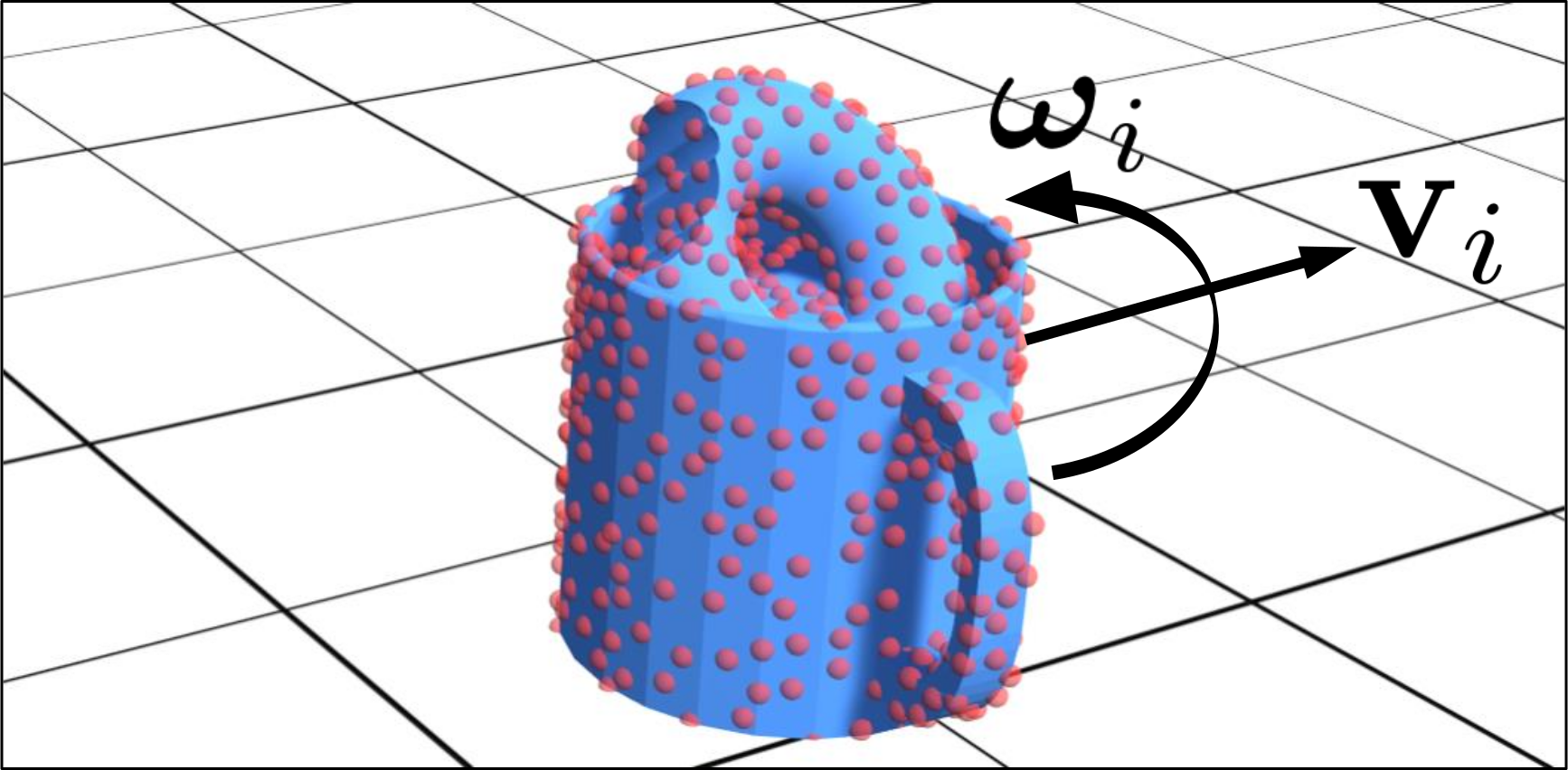}
\end{center}
   \caption{Problem input. Our method uses a point cloud (red spheres) and initial linear $\vect{v}_i$ (arrow) and angular velocity $\vect{\omega}_i$ (circular arrow) to predict dynamics. The object is assumed to move on a plane but can exhibit wobbling or complete toppling.
   }
\label{fig:formulation_detail}
\end{figure}

A single datapoint in each of our datasets is a unique simulation which begins with an object on a flat plane that has just been subjected to a random 3D impulsive force. This force results in the object acquiring initial linear and angular velocities and eventually comes to rest due to friction or topples over. We record the full state of the object (3D shape, 3D position, rotation, linear and angular velocities, and stability) at each discrete time step during the entire simulation. We use this information to derive the change in object state at each time step to train our network. An object is considered to be toppling (\ie the binary flag is set to unstable) when the angle between its up-axis and the global up-axis is greater than 45 degrees. Although we apply an initial impulsive force, we do not use this information in our network. This makes our method generally applicable with only the knowledge of initial velocities which could be estimated from point cloud observations or video.

\parahead{Simulation Procedure}
For our input, we use the same technique as \cite{qi2017pointnet} to sample a point cloud with 1024 points from the surface of each unique object in all datasets (see Figure~\ref{fig:formulation_detail}). ShapeNet objects are normalized to fit within a unit cube so the extents of the objects are about 0.5 m.
The applied impulsive force direction, magnitude, and position are chosen randomly from a uniform distribution around the object center of mass.
This helps the simulations span both sliding and toppling examples, and imparts both initial linear and angular velocities. Objects may translate up to 10 m and complete more than 10 complete rotations throughout the simulated trajectories. Simulations are usually 3 to 5 seconds long with data collected at 15 Hz. 
Friction coefficients and object density are the same across all simulations.
We use the exact mesh to build a collider that captures the object complexity during ground contact in simulation.

\parahead{Datasets}
We synthesize multiple categories of datasets to train and evaluate our models: \texttt{Primitives}, \texttt{Bottles}, \texttt{Mugs}, \texttt{Trashcans}, \texttt{Speakers}, and \texttt{Combined}. %
Training objects are simulated with a different random scale from 0.5 to 1.5 for x, y, and z directions in order to increase shape diversity. The \texttt{Primitives} dataset is further divided into a \texttt{Box} dataset which is a single cube scaled to various non-uniform dimensions, and a \texttt{Cylinders} dataset that contains a variety of cylinders.
The remaining four datasets represent everyday shape categories taken from the ShapeNet~\cite{chang2015shapenet} repository. These exhibit wide shape diversity and offer a more challenging task.
Lastly, we have a dataset which combines all of the objects and simulations from the previous six to create a large and diverse set of shapes which is split roughly evenly between categories. In total, we use \textbf{793} distinct object shapes and run \textbf{65,715} simulations to generate our data.

\begin{figure*}[!th]
\begin{center}
   \includegraphics[width=\textwidth]{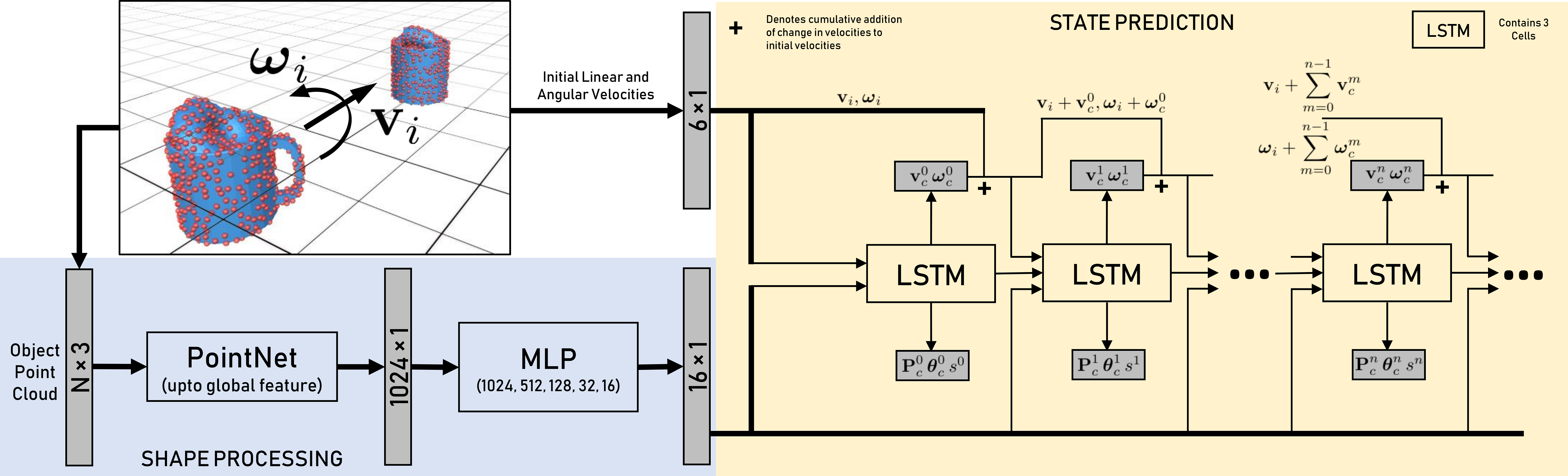}
\end{center}
   \caption{Model architecture. Our network takes the initial linear and angular velocities, and the object point cloud as input and predicts the change in the object's 3D position, rotation, linear and angular velocities, and object stability.
   The shape processing branch extracts shape features which are concatenated with the input velocities and fed to an LSTM (shown unrolled here) which makes the state change prediction at each time step. The input velocities are the cumulative sum of the estimated velocity changes and the initial velocities. Numbers in bracket indicate the output size of each layer and MLP indicates multilayer perceptron. 
}
\label{fig:arch}
\end{figure*}

%% file: content/05_method.tex
\section{Method}
\label{sec:method}
A straightforward approach to predict changes in object state could be to combine all inputs into one vector and use a neural network to directly predict state change at each step in a recursive fashion. This approach cannot learn the intricacies of object shape and non-linear object motion since it does not keep track of past states of the object.
We therefore use a combination of PointNet~\cite{qi2017pointnet}, which extracts shape features, and a recurrent neural network (RNN) which encodes past states and predicts future states of the object.

\subsection{Network Architecture}
The motion of an object throughout a trajectory depends on: (1)~the shape which affects mass $m$, moment of inertia $I$, and contact surface, and (2)~initial linear and angular velocities. We therefore design our network to learn important information related to the shape and initial velocities. Our model (see Figure~\ref{fig:arch}) is composed of two parts, a one-time shape processing branch and a state prediction branch.

\parahead{Shape Processing}
The shape processing branch is designed to extract salient shape features that are crucial to making accurate predictions. Object geometry affects both linear and angular velocities through its mass (which depends on volume) and moment of inertia. The aim of this branch is to help the network develop notions of volume, mass, and inertia from a point cloud. It must also learn the effect of the area and shape of the bottom contacting surface which determines how friction affects rotation. To this end, we use PointNet~\cite{qi2017pointnet}. As shown in Figure~\ref{fig:arch}, the initial object point cloud is fed to the PointNet classification network which outputs a global feature that is further processed to output a final \emph{shape feature}. %
Since the shape of rigid objects does not change, we extract a shape feature once during the first step and re-use it in subsequent steps.

\parahead{State Prediction}
The goal of the state prediction branch is to estimate the change in object state at each time step in a sequence.
Similar to other sequential problems~\cite{sutskever2011generating}, we use a recurrent neural network, and particularly a long short-term memory (LSTM) network to capture the temporal relationships in object state changes.
The input to our LSTM, which maintains a hidden state of size 1024, consists of a 22-dimensional vector which concatenates the initial linear and angular velocities, and the features extracted by the shape processing branch (see Figure~\ref{fig:arch}).
The LSTM predicts the \emph{change in object state}, \ie, change in 3D position, rotation, object stability ($\vect{P}_c, \vect{\theta}_c, s_c$), and linear and angular velocities ($\vect{v}_c, \vect{\omega}_c$). At test time, we would like to roll out an entire trajectory prediction. %
To do this, the input to the first step is the observed initial velocities ($\vect{v}_i, \vect{\omega}_i$). Then the LSTM-predicted change in velocity is summed with the input to arrive at the new object velocity (which is used as input to the subsequent step). This is performed recurrently to produce a full trajectory of relative positions and rotations, given only ground truth initial state.

\subsection{Loss Functions \& Training}
The goal of the network is to minimize the error between the predicted and ground truth state change.
We found that using $L^p$ losses for position, rotation, and velocities caused the network to focus too much on examples with large error. Instead we propose a form of relative error. For instance, for change in 3D position we use a relative $L^2$ error between the predicted position $\hat{\vect{P}}_c$ and the ground truth $\vect{P}_c$. We sum the values in the denominator to avoid numerical instability when ground truth change in position is near zero. Furthermore, we found that different components of the object state change required different losses for best performance. We use the $L^2$ loss for 3D position and linear and angular velocities. For rotation represented in axis-angle form, we use an $L^1$ loss. For change in 3D position and rotation:
\begin{align}
    \mathcal{L}_{\vect{P}} &= \frac{||\hat{\vect{P}}_c - \vect{P}_c||_2}{||\hat{\vect{P}}_c||_2 + ||\vect{P}_c||_2}, &
    \mathcal{L}_{\vect{\theta}} &= \frac{||\hat{\vect{\theta}}_c - \vect{\theta}_c||_1}{||\hat{\vect{\theta}}_c||_1 + ||\vect{\theta}_c||_1}.
    \label{eqn:loss}
\end{align}
We use binary cross entropy loss $\mathcal{L}_s$ for object stability. The losses for change in velocities are identical to that of position. Our final loss is the sum of $\mathcal{L} =  w_{\vect{P}}\mathcal{L}_{\vect{P}} + w_{\vect{\theta}}\mathcal{L}_{\vect{\theta}} + w_{\vect{v}}\mathcal{L}_{\vect{v}} + w_{\vect{\omega}}\mathcal{L}_{\vect{\omega}} + w_{s}\mathcal{L}_{s}$. Empirically, all objective weights are 1 except the stability term $w_{s} = 2$.

We train the state prediction LSTM on sequences of 15 timesteps (corresponding to 1 second of simulation). Each sequence is a random window chosen from simulations in the dataset. The loss is applied at every timestep. We train all branches of our network jointly using the Adam~\cite{kingma2015adam} optimization algorithm. %
In the shape processing branch, PointNet weights are pretrained on ModelNet40~\cite{wu2015modelnet}, then fine-tuned during our training process. %
Before training, 10\% of the objects in the training split are set aside as validation data for early stopping. %

%% file: content/06_results.tex
\input{content/06_01_summary_table.tex}
\section{Experiments}
\label{sec:results}
We present extensive experimental evaluation on the generalization ability of our method, compare to baselines and prior work, and show results on real-world data. We \emph{highly encourage} the reader to watch the supplementary video which gives a better idea of our data, along with the accuracy of predicted trajectories from the model.

\begin{figure*}[t]
\begin{center}
   \includegraphics[width=0.49\textwidth]{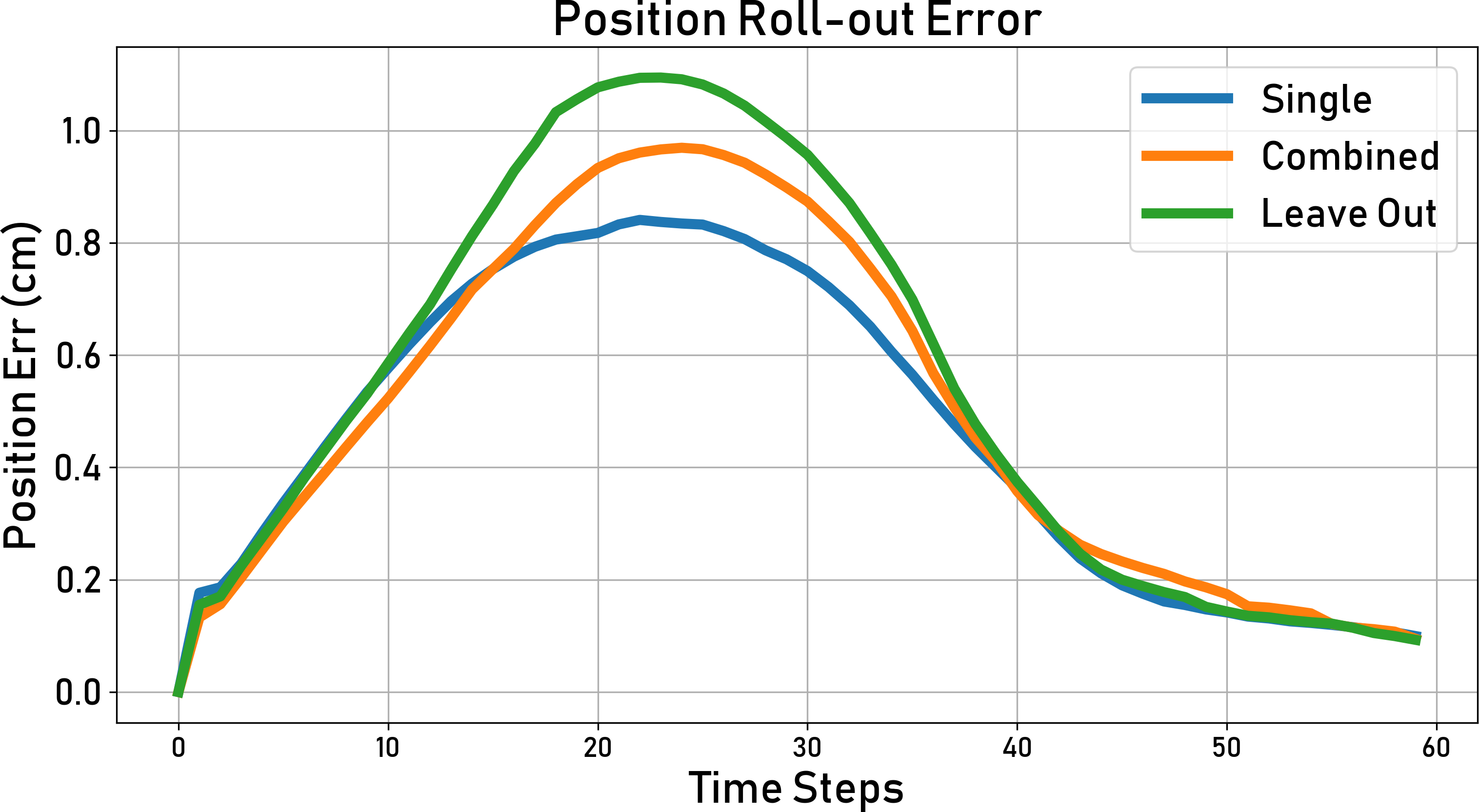}
   \includegraphics[width=0.49\textwidth]{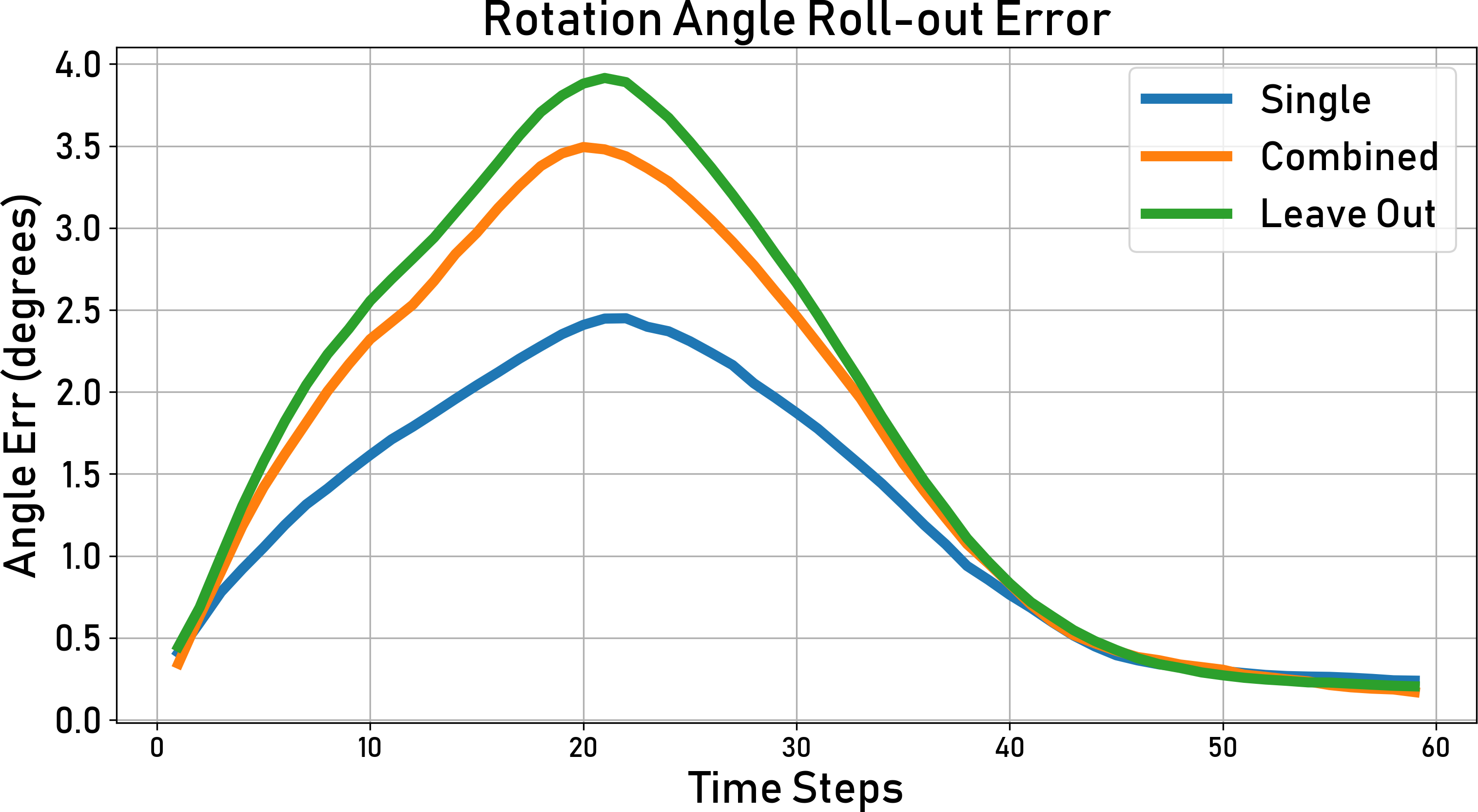}
\end{center}
   \caption{Roll-out errors for object generalization experiments. Each curve shows the median roll-out error over all evaluation datasets using that training procedure. Separate models trained on each dataset are shown by the \textcolor{blue}{blue curves}, a single model trained on the \texttt{Combined} dataset then evaluated on individual datasets is shown by the \textcolor{orange}{orange curve}, and separate models trained on the \texttt{Combined} dataset with the evaluation shape category left out are shown in \textcolor{green}{green}.}
\label{fig:rolloutresults}
\end{figure*}

\parahead{Evaluation Metrics}
For all experiments, we report both \textit{single-step} and \textit{roll-out} errors for dynamics predictions. Both errors measure the mean difference between the model's change in state prediction and ground truth over all timesteps in all test examples. The metrics differ due to the input used at each time step. \textit{Single-step} error uses the ground truth velocities as input to every timestep (the same process used in training). Single-step errors are shown in Table~\ref{table:resultssummary} for linear (cm/s) and angular (rad/s) velocity, position (cm), angle (deg), and rotation axis (measured as $1 - \cos{\alpha}$ where $\alpha$ is the angle between the predicted and ground truth axes). Single-step errors are reported for all test sequences, \textbf{including} those with toppling. On the other hand, \emph{roll-out} error measures the model's capability to roll out object trajectories given only the initial conditions. In this case, the network uses its own velocity predictions as input to each following step as described in Section~\ref{sec:method}. Roll-out errors for various models are shown in Figure~\ref{fig:rolloutresults}. Unless noted otherwise, reported roll-out errors are only for test sequences that \textbf{do not contain} toppling. This is done to focus evaluation on shape generalization without the stochasticity of toppling (see discussion in supplementary material). %

\subsection{Object Generalization}
We first perform object generalization experiments to evaluate whether the learned model is able to generalize to unseen objects---a crucial ability for autonomous systems in unseen environments. Since it is impossible to experience all objects that an agent will interact with, we would like knowledge of similarly-shaped objects to inform reasonable predictions about dynamics in new settings. For these experiments, we split datasets based on unique objects such that \textbf{no test objects are seen during training}. %
Since our network is designed specifically to process object shape and learn relevant physical properties, we expect it to extract general features allowing for accurate predictions even on novel objects. We evaluate models trained on both single and combined categories; all \emph{single-step} errors are shown in Table~\ref{table:resultssummary} and \emph{roll-out} errors in Figure~\ref{fig:rolloutresults}. 

\parahead{Single Category}
We train a separate network for each object category. Results for single-step errors on each dataset are shown in Table~\ref{table:resultssummary} under the procedure \emph{Single}, and roll-out errors over all evaluation datasets are shown by the \textcolor{blue}{blue curves} in Figure~\ref{fig:rolloutresults}. Our model makes accurate single-step predictions (with ground truth velocity input at each step) and is able to stay under 1 cm and 2.5 degrees error for position and rotation for unseen objects during roll out (using its own velocity predictions as input to each step). This indicates that the network is able to generalize to unseen objects within the same shape category.

\parahead{Combined Categories}
Next, our model is trained on the \texttt{Combined} dataset and then evaluated on all individual datasets. Single-step errors are shown under the \emph{Combined} training procedure in Table~\ref{table:resultssummary} and roll-out errors by the \textcolor{orange}{orange curve} in Figure~\ref{fig:rolloutresults}. In general, performance is very similar to training on individual datasets and even improves errors in many cases; for example, single-step errors on the \texttt{Mugs}, \texttt{Trashcans}, \texttt{Bottles}, and \texttt{Speakers}. This indicates that exposing the network to larger shape diversity at training time can help focus learning on underlying physical relationships rather than properties of a single group of objects. %
In order to maintain this high performance, the network is likely learning a general approach to extract salient physical features from the diverse objects in the \texttt{Combined} dataset rather than memorizing how specific shapes behave.

\begin{figure*}[t]
\begin{center}
   \includegraphics[width=0.99\textwidth]{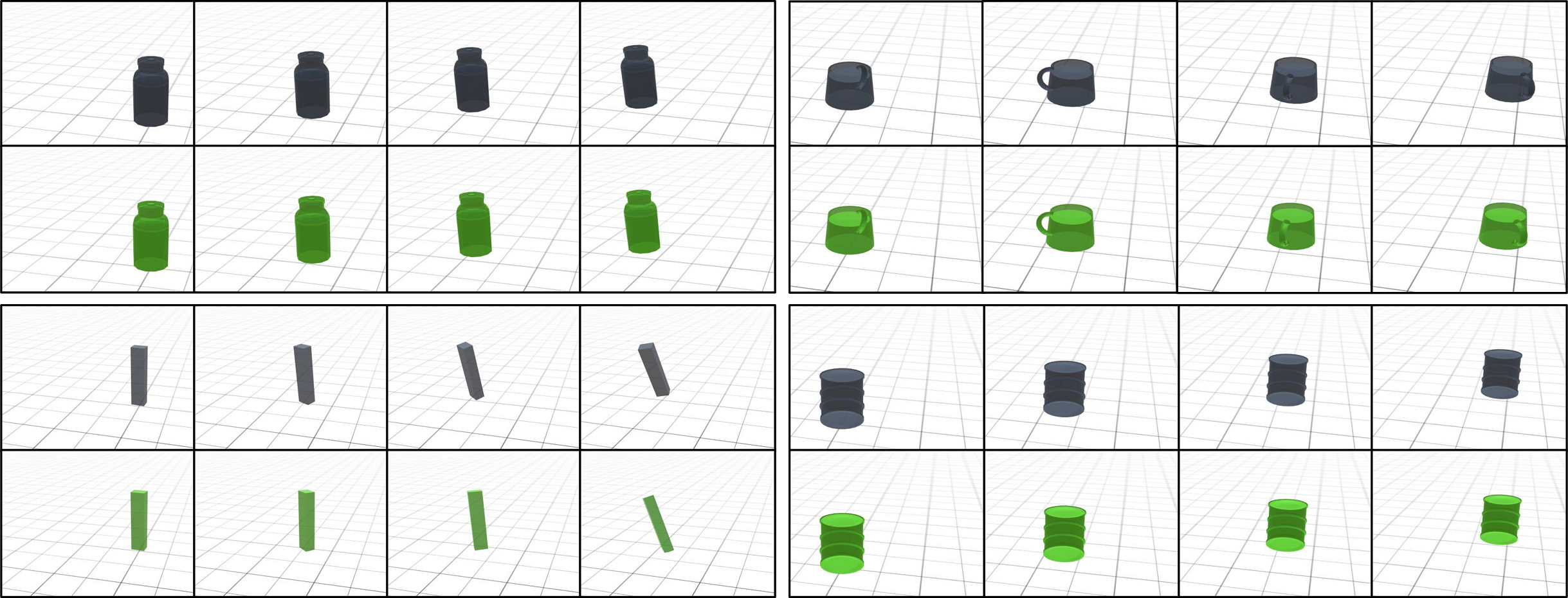}
\end{center}
   \caption{Qualitative results. Four sample frames from a sequence for models trained on the \texttt{Combined} dataset with the evaluation category left out. Ground truth simulation is shown in grey and the network-predicted trajectory in green. Three non-toppling examples are shown for \texttt{Bottles} (top left), \texttt{Mugs} (top right), and \texttt{Trashcans} (bottom right). A toppling result is shown for \texttt{Boxes} (bottom left).}
\label{fig:predresults}
\end{figure*} 

\parahead{Out of Category}
Lastly, we evaluate performance on the extreme task of generalizing \textbf{outside of trained object categories}. For this, we create new \texttt{Combined} datasets each with one object category left out of the training set. We then evaluate its performance on objects from the left out category. Single-step errors for these experiments are shown under the \emph{Leave Out} heading in Table~\ref{table:resultssummary} and roll-out errors appear in the \textcolor{green}{green curve} in Figure~\ref{fig:rolloutresults}. We see only a slight drop in single-step performance for almost every evaluation shape category. Additionally, mean roll-out errors reach less than 1.2 cm and 4 degrees for position and rotation angle, respectively. Overall, this result shows the model can make accurate predictions for objects from completely different categories in spite of their dissimilarity to training shapes.
The model seems to have developed a deep understanding of how shape affects dynamics through mass, moment of inertia, and contact surface in order to generalize to novel categories. Some trajectories from leave-one-out trained models are visualized in Figure~\ref{fig:predresults}. %

\parahead{Toppling Classification}
\label{sec:toppling_class}
In addition to predicting object state, our model also classifies whether the object is currently toppling at each time step (the binary stability flag). %
Here we evaluate the ability to classify entire trajectories as toppling or not. To do this, we consider a rolled-out trajectory as toppling if the model predicts the object is unstable for any step in the rolled-out sequence. We find that a single model trained on the \texttt{Combined} dataset is able to achieve an average F-score of 0.64 on this classification task for the \texttt{Boxes}, \texttt{Cylinders}, and \texttt{Bottles} datasets. Roughly half of the simulations in these datasets contain toppling (see supplement), so the model has sufficient examples to learn what features of motion indicate probable instability.

\subsection{Friction Generalization}
\label{sec:frictiongen}
One advantage of learning dynamics over traditional simulation is the ability to implicitly represent physical properties of a system. Our LSTM achieves this by aggregating information in its hidden state. This is exemplified in the ability to adapt to unknown friction coefficients at test time. In this experiment, we create a new \texttt{Speakers} dataset where the object in each simulation has a randomly chosen friction coefficient from a uniform distribution between 0.35 and 0.65. We train our model on this new dataset, and compare its ability to roll out trajectories against the model trained on constant-friction data. Roll-out errors are shown in Table~\ref{table:frictionresults}. Unsurprisingly, the model trained on the varied friction data is less accurate than the constant model given only initial velocities. With only initial conditions, there is no way for the model to infer the object friction. Therefore, we allow the varied friction model to use additional ground truth velocity steps at the beginning of its test-time roll out (indicated by ``Steps In" in Table~\ref{table:frictionresults}), which allows it to implicitly infer the friction using the LSTM's hidden state. As seen in Table~\ref{table:frictionresults}, when the model trained on varied friction data uses 6 input steps, its performance is as good as the constant-friction model. This shows the model's ability to accurately generalize to new frictions if allowed to observe a small portion ($< 0.35$ seconds) of the object's motion.

\subsection{Comparison to MLP Baseline}
We justify the use of a memory mechanism by comparing our proposed model to a modified architecture where the LSTM in the state prediction branch is replaced with a simple MLP containing 5 fully-connected layers. We train and evaluate both models on the \texttt{Speakers} dataset (with constant friction). The baseline MLP architecture has no memory, so it predicts based on the velocities and shape feature at each step. This is a natural approach which assumes the future physical state of an object relies only on its current state. However, as shown in Table~\ref{table:mlpbaseline}, this model %
gives worse results, especially for position and angle. This may be because a hidden state gives the network some notion of acceleration over multiple timesteps and allows for self-correction during trajectory roll out.

\begin{table}[t]
\begin{center}
\scalebox{0.8}{
\begin{tabular}{l c l l l l l}
\toprule
\textbf{Data} & \textbf{Steps In} & $\vect{v}$ & $\vect{\omega}$ & $\vect{P}$ & $|\vect{\theta}|$ & $\bar{\vect{\theta}}$ \\
\midrule
Constant Friction & 1  & \textbf{1.993}	& 0.098 &	0.369	& 0.743 &	\textbf{0.016} \\ 
Vary Friction & 1  & 2.918	& 0.112 &	0.723	& 1.283 &	0.057 \\ 
Vary Friction & 4  & 2.287 &	0.098 &	0.417 &	0.674 &	0.033  \\
Vary Friction & 6  & 2.163 &	\textbf{0.094} &	\textbf{0.358} &	\textbf{0.575} &	0.029  \\
\bottomrule
\end{tabular}}
\end{center}
\caption{Roll-out errors (same units as Table~\ref{table:resultssummary}) for friction generalization experiments. Our model is trained on the \texttt{Speakers} dataset with constant a friction coefficient of 0.5 and with friction randomly varied from 0.35 to 0.65. Test-time roll-outs use a varied number of observed velocity input steps (\emph{Steps In}).}
\label{table:frictionresults}
\end{table}

\begin{table}[t]
\begin{center}
\scalebox{0.8}{
\begin{tabular}{l l l l l l}
\toprule
\textbf{State Predictor} & $\vect{v}$ & $\vect{\omega}$ & $\vect{P}$ & $|\vect{\theta}|$ & $\bar{\vect{\theta}}$ \\
\midrule
LSTM & \textbf{1.786} &	\textbf{0.112} &	\textbf{0.096} &	\textbf{0.233}	& \textbf{0.044} \\ 
MLP & 2.770  & 0.194	& 0.286 &	0.819	& 0.061 \\
\bottomrule
\end{tabular}}
\end{center}
\caption{Single-step errors (same units as Table~\ref{table:resultssummary}) training on the \texttt{Speakers} dataset with our proposed state predictor (LSTM) against an MLP baseline with no memory.}
\label{table:mlpbaseline}
\end{table}

\subsection{Comparison to Other Work}
We compare our method to the \emph{hierarchical relation network} (HRN)~\cite{mrowca2018flexible} to highlight the differences between an object-centric (our work) approach and their particle-based method. Both models are trained on a small dataset of 1519 scaled boxes simulated in the NVIDIA FleX engine~\cite{macklin2014unified}, then evaluated on 160 held out simulations. Each simulation contains a box sliding with some initial velocity which comes to rest without toppling. We compare the mean roll-out errors of the two models. Our model averages \textbf{0.51} cm and \textbf{0.36} degrees \emph{roll-out} errors for position and rotation angle, respectively, while HRN achieves \textbf{1.99} cm and \textbf{2.73} degrees. An object-centric approach seems to simplify the job of the prediction network offering improved accuracy over individually predicting trajectories of particles that make up a rigid object. We note, however, that HRN shows prediction ability on falling rigid objects and deformables, which our model can not handle.

\subsection{Real-World Data}
\label{sec:real_results}
\begin{figure}[t]
\begin{center}
   \includegraphics[width=0.99\linewidth]{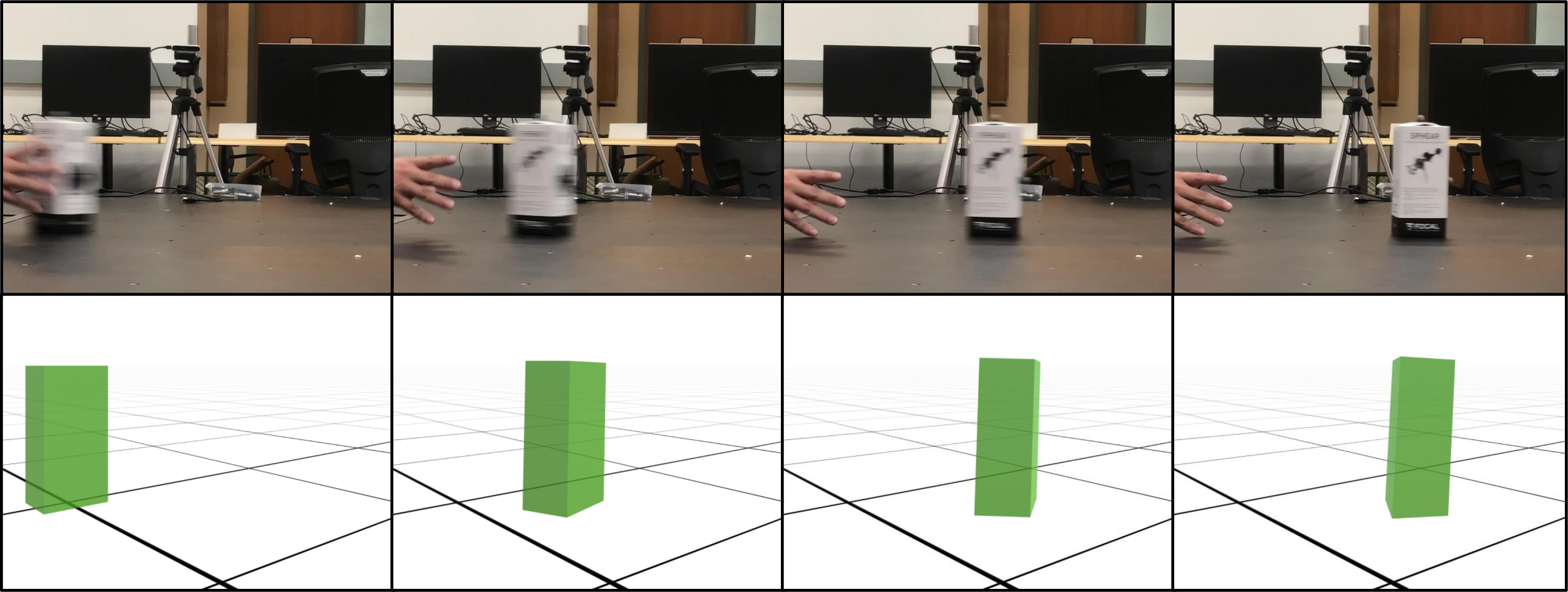}
\end{center}
   \caption{Real-world data. We captured 66 sequences of a box with a motion capture system and trained our method on the captured data. The top row shows an external view of one of the test trials. The bottom row shows predictions.}
\label{fig:real}
\end{figure}

To show our model's ability to generalize to the real world, we captured 66 trials of a small sliding box using a motion capture system which provides full object state information throughout a trajectory. From this we extract all necessary training data then construct a point cloud based on the box measurements. We train our model \textbf{directly on} 56 of the trials and test on 10 held-out trajectories. For real-world data, we give the model 2 steps of initial velocity input, which we found improved performance. Similar to the friction experiments in Section~\ref{sec:frictiongen}, having multiple steps as input allows the network to perform implicit parameter identification since the real-world data is much noisier than in simulation (\ie different parts of the table may have slightly different friction properties) and possibly helps the network identify initial acceleration. Despite the lack of data, our model is able to reasonably learn the complex real-world dynamics achieving %
single-step errors of 8.3 cm/s,  0.733 rad/s,  0.289 cm, 1.13 degrees, and 0.436 (for axis). We visualize a predicted trajectory in Figure~\ref{fig:real}.

%% file: content/06_01_summary_table.tex
\begin{table*}[!ht]
\begin{center}
\scalebox{0.84}{
\begin{tabular}{l l l l l l l | l l l l l l l}
\toprule
\textbf{Test Set} & \textbf{Procedure} & $\vect{v}$ & $\vect{\omega}$ & $\vect{P}$ & $|\vect{\theta}|$ & $\bar{\vect{\theta}}$ & \textbf{Test Set} & \textbf{Procedure} & $\vect{v}$ & $\vect{\omega}$ & $\vect{P}$ & $|\vect{\theta}|$ & $\bar{\vect{\theta}}$\\
\midrule
\texttt{Box} & Single  & \textbf{2.615} &	\textbf{0.201}	& 0.111 &	0.460 &	0.148 & \texttt{Trashcans} & Single  & 3.014	& 0.168 &	0.144 &	0.247 &	0.040 \\ 
 & Combined  & 2.696 &	0.209	 & \textbf{0.107}	& \textbf{0.453} &	\textbf{0.140} & & Combined  &  \textbf{2.858} &	\textbf{0.162} &	\textbf{0.138} &	\textbf{0.226} &	\textbf{0.032}\\ 
 & Leave Out  & 2.661 &	0.208	& 0.107 &	0.454 &	0.161
 & & Leave Out  & 2.918 &	0.165 &	0.142 &	0.237 &	0.035 \\ 
\midrule
\texttt{Cylinders} & Single  & \textbf{4.235}	& \textbf{0.228} & \textbf{0.152} &	\textbf{0.489} &	0.029 & \texttt{Bottles} & Single  & 4.894 &	0.264 &	0.654 &	0.993 &	\textbf{0.029} \\ 
 & Combined & 4.597	& 0.238 &	0.157 &	0.492 &	0.030 & & Combined  & \textbf{4.662} &	\textbf{0.247}	& \textbf{0.652} &	\textbf{0.992} & 0.030 \\ 
 & Leave Out  & 4.851 &	0.255 &	0.165 &	0.518 &	\textbf{0.024} & & Leave Out  & 4.891	& 0.264	 & 0.658 &	1.010 &	0.029  \\ 
\midrule
\texttt{Mugs} & Single  & 2.851 &	0.179 &	0.113 &	0.207 &	0.019 & \texttt{Speakers} & Single  & 1.786 &	0.112 &	0.096 &	0.233	& 0.044\\ 
 & Combined  & \textbf{2.723} &	\textbf{0.173} &	\textbf{0.099} &	\textbf{0.181} &	0.019 &  & Combined  & \textbf{1.675} &	\textbf{0.106} &	\textbf{0.082} &	\textbf{0.200}	& \textbf{0.040} \\ 
 & Leave Out  & 2.781	& 0.177 &	0.104 &	0.198 &	\textbf{0.018}  & & Leave Out  & 1.770 &	0.110 &	0.084 &	0.223 &	0.048 \\
\midrule
\texttt{Combined} & Combined  & 3.175 &	0.184 &	0.218 &	0.417 &	0.041 \\ 
\bottomrule
\end{tabular}}
\end{center}
\caption{Single-step errors for object generalization experiments. For each dataset, we show the single-step evaluation errors when a model is trained on that \emph{Single} dataset, the \texttt{Combined} dataset which contains all shape categories, and the \texttt{Combined} dataset with the evaluation category \emph{left out}. Errors are in cm/s for linear velocity $\vect{v}$, rad/s for angular velocity $\vect{\omega}$, cm for position $\vect{P}$, degrees for rotation angle $|\vect{\theta}|$, and 1 - $\cos\alpha$ for axis $\bar{\vect{\theta}}$. Single-step errors are the mean difference between predicted change in state and ground truth change given the ground truth as input to each step.}
\label{table:resultssummary}
\end{table*}

%% file: content/07_limitations.tex
\section{Limitations and Future Work}
\label{sec:limitations}
Our approach has limitations and there remains room for future work.
Here we focused on learning the 3D dynamics of objects on a planar surface by capturing sliding dynamics.
However, free 3D dynamics and complex phenomena such as collisions are not captured in our work and presents important directions for future work. Additionally, we avoid the uncertainty inherent to toppling in favor of evaluating shape generalization, but capturing this stochasticity is important for future work. We believe that our approach provides a strong foundation for developing methods for these complex motions.
Our method is fully supervised and does not explicitly model physical laws like some previous work~\cite{stewart2017label}. We show some examples on real-world data but more complex motion from camera-based sensing is a topic for future work and will be important to exploit the potential for our learned model to improve accuracy over analytic simulation.
Results from Section~\ref{sec:frictiongen} indicate our model's potential for physical parameter estimation, but we largely ignore this problem in the current work by assuming constant friction and density for most experiments.

%% file: content/08_conclusion.tex
\section{Conclusion}
We presented a method for learning to predict the 3D physical dynamics of a rigid object moving along a plane with an initial velocity. Our method is capable of generalizing to previously unseen object shapes and new initial velocities not seen during training. We showed that this challenging dynamics prediction problem can be solved using a neural network architecture that is informed by physical laws. We train our network on 3D point clouds of a large shape collection and a large synthetic dataset with experiments showing that we are able to accurately predict the change in state for sliding objects. We additionally show the model's ability to learn directly from real-world data.

%% file: LearningDynamics_WACV2020.bbl
\begin{thebibliography}{10}\itemsep=-1pt

\bibitem{bullet}
Bullet physics engine.
\newblock \url{https://pybullet.org}.

\bibitem{unity}
Unity game engine.
\newblock \url{https://unity3d.com}.

\bibitem{agrawal2016poke}
P.~Agrawal, A.~Nair, P.~Abbeel, J.~Malik, and S.~Levine.
\newblock Learning to poke by poking: Experiential learning of intuitive
  physics.
\newblock In {\em Proceedings of the 30th Conference on Neural Information
  Processing Systems (NIPS)}, 2016.

\bibitem{ajay2018pushing}
A.~Ajay, J.~Wu, N.~Fazeli, M.~Bauz{\'{a}}, L.~P. Kaelbling, J.~B. Tenenbaum,
  and A.~Rodriguez.
\newblock Augmenting physical simulators with stochastic neural networks: Case
  study of planar pushing and bouncing.
\newblock In {\em International Conference on Intelligent Robots and Systems
  (IROS)}, 2018.

\bibitem{baillargeon1990top}
R.~Baillargeon and S.~Hanko-Summers.
\newblock Is the top object adequately supported by the bottom object? young
  infants' understanding of support relations.
\newblock {\em Cognitive Development}, 5(1):29--53, 1990.

\bibitem{battaglia2016interactionnets}
P.~Battaglia, R.~Pascanu, M.~Lai, D.~J. Rezende, and K.~kavukcuoglu.
\newblock Interaction networks for learning about objects, relations and
  physics.
\newblock In {\em Proceedings of the 30th International Conference on Neural
  Information Processing Systems (NIPS)}, pages 4509--4517, 2016.

\bibitem{battaglia2013simunderstanding}
P.~W. Battaglia, J.~B. Hamrick, and J.~B. Tenenbaum.
\newblock Simulation as an engine of physical scene understanding.
\newblock {\em Proceedings of the National Academy of Sciences},
  110(45):18327--18332, 2013.

\bibitem{byravan2017se3}
A.~Byravan and D.~Fox.
\newblock Se3-nets: Learning rigid body motion using deep neural networks.
\newblock In {\em 2017 IEEE International Conference on Robotics and Automation
  (ICRA)}, 2017.

\bibitem{byravan2018se3pose}
A.~Byravan, F.~Leeb, F.~Meier, and D.~Fox.
\newblock Se3-pose-nets: Structured deep dynamics models for visuomotor
  planning and control.
\newblock In {\em IEEE International Conference on Robotics and Automation
  (ICRA)}, 2018.

\bibitem{chang2015shapenet}
A.~X. Chang, T.~Funkhouser, L.~Guibas, P.~Hanrahan, Q.~Huang, Z.~Li,
  S.~Savarese, M.~Savva, S.~Song, H.~Su, et~al.
\newblock Shapenet: An information-rich 3d model repository.
\newblock {\em arXiv preprint arXiv:1512.03012}, 2015.

\bibitem{chang2017compositional}
M.~B. Chang, T.~Ullman, A.~Torralba, and J.~B. Tenenbaum.
\newblock A compositional object-based approach to learning physical dynamics.
\newblock In {\em Proceedings of the 5th International Conference on Learning
  Representations (ICLR)}, 2017.

\bibitem{peres2018diffengine}
F.~de~Avila Belbute-Peres, K.~Smith, K.~Allen, J.~Tenenbaum, and J.~Z. Kolter.
\newblock End-to-end differentiable physics for learning and control.
\newblock In {\em Advances in Neural Information Processing Systems (NeurIPS)},
  2018.

\bibitem{ehrhardt2017longterm}
S.~Ehrhardt, A.~Monszpart, N.~{J. Mitra}, and A.~Vedaldi.
\newblock {Learning A Physical Long-term Predictor}.
\newblock {\em arXiv preprint, arXiv:1703.00247}, Mar. 2017.

\bibitem{ehrhardt2018visualobs}
S.~Ehrhardt, A.~Monszpart, N.~J. Mitra, and A.~Vedaldi.
\newblock Unsupervised intuitive physics from visual observations.
\newblock {\em arXiv preprint, arXiv:1805.05086}, 2018.

\bibitem{ehrhardt2017mechanics}
S.~Ehrhardt, A.~Monszpart, A.~Vedaldi, and N.~{J. Mitra}.
\newblock {Learning to Represent Mechanics via Long-term Extrapolation and
  Interpolation}.
\newblock {\em arXiv preprint arXiv:1706.02179}, June 2017.

\bibitem{finn2016videoprediction}
C.~Finn, I.~Goodfellow, and S.~Levine.
\newblock Unsupervised learning for physical interaction through video
  prediction.
\newblock In {\em Proceedings of the 30th International Conference on Neural
  Information Processing Systems (NIPS)}, pages 64--72, 2016.

\bibitem{finn2017planning}
C.~Finn and S.~Levine.
\newblock Deep visual foresight for planning robot motion.
\newblock In {\em International Conference on Robotics and Automation (ICRA)},
  2017.

\bibitem{fraccaro2017disentangled}
M.~Fraccaro, S.~Kamronn, U.~Paquet, and O.~Winther.
\newblock A disentangled recognition and nonlinear dynamics model for
  unsupervised learning.
\newblock In {\em Advances in Neural Information Processing Systems (NIPS)},
  2017.

\bibitem{fragkiadaki2016visualbilliards}
K.~Fragkiadaki, P.~Agrawal, S.~Levine, and J.~Malik.
\newblock Learning visual predictive models of physics for playing billiards.
\newblock In {\em Proceedings of the 4th International Conference on Learning
  Representations (ICLR)}, 2016.

\bibitem{grzeszczuk2000neuroanimator}
R.~Grzeszczuk, D.~Terzopoulos, and G.~Hinton.
\newblock {\em NeuroAnimator: fast neural network emulation and control of
  physics-based models.}
\newblock University of Toronto, 2000.

\bibitem{hamrick2016decision}
J.~B. Hamrick, R.~Pascanu, O.~Vinyals, A.~Ballard, N.~Heess, and P.~Battaglia.
\newblock Imagination-based decision making with physical models in deep neural
  networks.
\newblock In {\em Advances in Neural Information Processing Systems (NIPS),
  Intuitive Physics Workshop}, 2016.

\bibitem{hu2019chainqueen}
Y.~Hu, J.~Liu, A.~Spielberg, J.~B. Tenenbaum, W.~T. Freeman, J.~Wu, D.~Rus, and
  W.~Matusik.
\newblock Chainqueen: {A} real-time differentiable physical simulator for soft
  robotics.
\newblock In {\em IEEE International Conference on Robotics and Automation
  (ICRA)}, 2019.

\bibitem{janner2019ooprediction}
M.~Janner, S.~Levine, W.~T. Freeman, J.~B. Tenenbaum, C.~Finn, and J.~Wu.
\newblock Reasoning about physical interactions with object-oriented prediction
  and planning.
\newblock In {\em International Conference on Learning Representations (ICLR)},
  2019.

\bibitem{kingma2015adam}
D.~P. Kingma and J.~Ba.
\newblock Adam: {A} method for stochastic optimization.
\newblock In {\em International Conference for Learning Representations
  (ICLR)}, 2015.

\bibitem{kipf2018neural}
T.~Kipf, E.~Fetaya, K.-C. Wang, M.~Welling, and R.~Zemel.
\newblock Neural relational inference for interacting systems.
\newblock {\em International Conference on Machine Learning (ICML)}, 2018.

\bibitem{kubricht2017intuitive}
J.~R. Kubricht, K.~J. Holyoak, and H.~Lu.
\newblock Intuitive physics: Current research and controversies.
\newblock {\em Trends in cognitive sciences}, 21(10):749--759, 2017.

\bibitem{lerer2016fbtowers}
A.~Lerer, S.~Gross, and R.~Fergus.
\newblock Learning physical intuition of block towers by example.
\newblock In {\em Proceedings of the 33rd International Conference on
  International Conference on Machine Learning (ICML)}, pages 430--438, 2016.

\bibitem{leslie1982perception}
A.~M. Leslie.
\newblock The perception of causality in infants.
\newblock {\em Perception}, 11(2):173--186, 1982.

\bibitem{li2016fall}
W.~Li, S.~Azimi, A.~Leonardis, and M.~Fritz.
\newblock To fall or not to fall: {A} visual approach to physical stability
  prediction.
\newblock {\em arXiv preprint, arXiv:1604.00066}, 2016.

\bibitem{li2017stability}
W.~Li, A.~Leonardis, and M.~Fritz.
\newblock Visual stability prediction for robotic manipulation.
\newblock In {\em 2017 IEEE International Conference on Robotics and Automation
  (ICRA)}, pages 2606--2613, May 2017.

\bibitem{li2019particledynamics}
Y.~Li, J.~Wu, R.~Tedrake, J.~B. Tenenbaum, and A.~Torralba.
\newblock Learning particle dynamics for manipulating rigid bodies, deformable
  objects, and fluids.
\newblock In {\em International Conference on Learning Representations (ICLR)},
  2019.

\bibitem{li2019propnet}
Y.~Li, J.~Wu, J.~Zhu, J.~B. Tenenbaum, A.~Torralba, and R.~Tedrake.
\newblock Propagation networks for model-based control under partial
  observation.
\newblock In {\em IEEE International Conference on Robotics and Automation
  (ICRA)}, 2019.

\bibitem{liu2018ppd}
Z.~Liu, W.~T. Freeman, J.~B. Tenenbaum, and J.~Wu.
\newblock Physical primitive decomposition.
\newblock In {\em Proceedings of the 15th European Conference on Computer
  Vision (ECCV)}, 2018.

\bibitem{macklin2014unified}
M.~Macklin, M.~M{\"u}ller, N.~Chentanez, and T.-Y. Kim.
\newblock Unified particle physics for real-time applications.
\newblock {\em ACM Transactions on Graphics (TOG)}, 33(4):153, 2014.

\bibitem{mirza2016unsupervised}
M.~{Mirza}, A.~{Courville}, and Y.~{Bengio}.
\newblock {Generalizable Features From Unsupervised Learning}.
\newblock {\em arXiv preprint, arXiv:1612.03809}, 2016.

\bibitem{monszpart2016SMASH}
A.~Monszpart, N.~Thuerey, and N.~J. Mitra.
\newblock {SMASH: Physics-guided Reconstruction of Collisions from Videos}.
\newblock {\em {ACM} Trans. Graph. ({SIGGRAPH} Asia)}, 2016.

\bibitem{mottaghi2016newton}
R.~Mottaghi, H.~Bagherinezhad, M.~Rastegari, and A.~Farhadi.
\newblock Newtonian image understanding: Unfolding the dynamics of objects in
  static images.
\newblock In {\em Proc. Computer Vision and Pattern Recognition (CVPR)}, 2016.

\bibitem{mottaghi2016if}
R.~Mottaghi, M.~Rastegari, A.~Gupta, and A.~Farhadi.
\newblock ``what happens if..." learning to predict the effect of forces in
  images.
\newblock In {\em Proceedings the 14th European Conference on Computer Vision
  (ECCV)}, 2016.

\bibitem{mrowca2018flexible}
D.~Mrowca, C.~Zhuang, E.~Wang, N.~Haber, L.~Fei-Fei, J.~B. Tenenbaum, and
  D.~L.~K. Yamins.
\newblock Flexible neural representation for physics prediction.
\newblock In {\em Proceedings of the 32nd International Conference on Neural
  Information Processing Systems (NIPS)}, 2018.

\bibitem{oh2015atari}
J.~Oh, X.~Guo, H.~Lee, R.~L. Lewis, and S.~P. Singh.
\newblock Action-conditional video prediction using deep networks in atari
  games.
\newblock In {\em Advances in Neural Information Processing Systems (NIPS)},
  2015.

\bibitem{qi2017pointnet}
C.~R. Qi, H.~Su, K.~Mo, and L.~J. Guibas.
\newblock Pointnet: Deep learning on point sets for 3d classification and
  segmentation.
\newblock {\em Proc. Computer Vision and Pattern Recognition (CVPR), IEEE},
  1(2):4, 2017.

\bibitem{rempe2019finalstate}
D.~Rempe, S.~Sridhar, H.~Wang, and L.~J. Guibas.
\newblock Learning generalizable physical dynamics of 3d rigid objects.
\newblock {\em arXiv preprint, arXiv:1901.00466}, 2019.

\bibitem{riochet2018intphys}
R.~Riochet, M.~Y. Castro, M.~Bernard, A.~Lerer, R.~Fergus, V.~Izard, and
  E.~Dupoux.
\newblock Intphys: {A} framework and benchmark for visual intuitive physics
  reasoning.
\newblock {\em arXiv preprint, arXiv:1803.07616}, 2018.

\bibitem{sanchez2018graphnet}
A.~{Sanchez-Gonzalez}, N.~{Heess}, J.~T. {Springenberg}, J.~{Merel},
  M.~{Riedmiller}, R.~{Hadsell}, and P.~{Battaglia}.
\newblock {Graph networks as learnable physics engines for inference and
  control}.
\newblock In {\em Proceedings the 35th International Conference on Machine
  Learning (ICML)}, 2018.

\bibitem{schenck2018spnets}
C.~Schenck and D.~Fox.
\newblock Spnets: Differentiable fluid dynamics for deep neural networks.
\newblock In {\em Conference on Robot Learning (CoRL)}, 2018.

\bibitem{smith2019intphys}
K.~Smith, L.~Mei, S.~Yao, J.~Wu, E.~Spelke, J.~Tenenbaum, and T.~Ullman.
\newblock Modeling expectation violation in intuitive physics with coarse
  probabilistic object representations.
\newblock In {\em Advances in Neural Information Processing Systems (NeurIPS)},
  2019.

\bibitem{stewart2017label}
R.~Stewart and S.~Ermon.
\newblock Label-free supervision of neural networks with physics and domain
  knowledge.
\newblock In {\em Proc. of AAAI Conference on Artificial Intelligence}, 2017.

\bibitem{sutskever2011generating}
I.~Sutskever, J.~Martens, and G.~E. Hinton.
\newblock Generating text with recurrent neural networks.
\newblock In {\em Proceedings of the 28th International Conference on Machine
  Learning (ICML-11)}, pages 1017--1024, 2011.

\bibitem{steenkiste2018relationalem}
S.~van Steenkiste, M.~Chang, K.~Greff, and J.~Schmidhuber.
\newblock Relational neural expectation maximization: Unsupervised discovery of
  objects and their interactions.
\newblock In {\em International Conference on Learning Representations (ICLR)},
  2018.

\bibitem{wang2018physnet}
Z.~Wang, S.~Rosa, B.~Yang, S.~Wang, N.~Trigoni, and A.~Markham.
\newblock 3d-physnet: Learning the intuitive physics of non-rigid object
  deformations.
\newblock In {\em Proceedings of the 26th International Joint Conference on
  Artificial Intelligence, {IJCAI-18}}, pages 4958--4964, 2018.

\bibitem{watters2017vin}
N.~Watters, A.~Tacchetti, T.~Weber, R.~Pascanu, P.~Battaglia, and D.~Zoran.
\newblock Visual interaction networks.
\newblock {\em arXiv preprint, arXiv:1706.01433}, 2017.

\bibitem{wu2016phys101}
J.~Wu, J.~J. Lim, H.~Zhang, J.~B. Tenenbaum, and W.~T. Freeman.
\newblock Physics 101: Learning physical object properties from unlabeled
  videos.
\newblock In {\em Proceedings of the 27th British Machine Vision Conference
  (BMVC)}, 2016.

\bibitem{wu2017deanimation}
J.~Wu, E.~Lu, P.~Kohli, W.~T. Freeman, and J.~B. Tenenbaum.
\newblock Learning to see physics via visual de-animation.
\newblock In {\em Proceedings of the 31st Conference on Neural Information
  Processing Systems (NIPS)}, 2017.

\bibitem{wu2015galileo}
J.~Wu, I.~Yildirim, J.~J. Lim, W.~T. Freeman, and J.~B. Tenenbaum.
\newblock Galileo: Perceiving physical object properties by integrating a
  physics engine with deep learning.
\newblock In {\em Proceedings of the 29th Conference on Neural Information
  Processing Systems (NIPS)}, pages 127--135, 2015.

\bibitem{wu2015modelnet}
Z.~Wu, S.~Song, A.~Khosla, F.~Yu, L.~Zhang, X.~Tang, and J.~Xiao.
\newblock 3d shapenets: A deep representation for volumetric shapes.
\newblock In {\em Computer Vision and Pattern Recognition (CVPR)}, 2015.

\bibitem{xu2019densephysnet}
Z.~Xu, J.~Wu, A.~Zeng, J.~B. Tenenbaum, and S.~Song.
\newblock Densephysnet: Learning dense physical object representations via
  multi-step dynamic interactions.
\newblock In {\em Robotics: Science and Systems (RSS)}, 2019.

\bibitem{ye2018interpretable}
T.~Ye, X.~Wang, J.~Davidson, and A.~Gupta.
\newblock Interpretable intuitive physics model.
\newblock In {\em Proceedings the 15th European Conference on Computer Vision
  (ECCV)}, pages 89--105, 2018.

\bibitem{yu2016push}
K.-T. Yu, M.~Bauz{\'a}, N.~Fazeli, and A.~Rodriguez.
\newblock More than a million ways to be pushed. a high-fidelity experimental
  dataset of planar pushing.
\newblock {\em IROS}, 2016.

\bibitem{zhang2016blocks}
R.~Zhang, J.~Wu, C.~Zhang, W.~T. Freeman, and J.~B. Tenenbaum.
\newblock A comparative evaluation of approximate probabilistic simulation and
  deep neural networks as accounts of human physical scene understanding.
\newblock In {\em Annual Meeting of the Cognitive Science Society}, 2016.

\bibitem{zheng2018percpred}
D.~Zheng, V.~Luo, J.~Wu, and J.~B. Tenenbaum.
\newblock Unsupervised learning of latent physical properties using
  perception-prediction networks.
\newblock In {\em Conference on Uncertainty in Artificial Intelligence (UAI)},
  2018.

\end{thebibliography}
